\documentclass{llncs}
\usepackage{graphicx}
\usepackage{amsmath,amssymb}
\usepackage{algorithmic}
\usepackage{xcolor}
\usepackage{soul} 
\usepackage{booktabs}
\usepackage{placeins}  

\begin{document}

\title{Gated-Attention Feature-Fusion Based Framework for Poverty Prediction}
\titlerunning{Poverty Assessment via Satellite Imagery}

\author{
Muhammad Umer Ramzan\textsuperscript{1} \and
Wahab Khaddim\textsuperscript{1} \and
Muhammad Ehsan Rana\textsuperscript{2} \and
Usman Ali\textsuperscript{1} \and
Manohar Ali\textsuperscript{1} \and
Fiaz ul Hassan\textsuperscript{1} \and
Fatima Mehmood\textsuperscript{1}
}

\authorrunning{M.U. Ramzan et al.}

\institute{
\textsuperscript{1}Department of Data Science, GIFT University, Gujranwala, Pakistan\\
\textsuperscript{2}Asia Pacific University of Technology and Innovation, Malaysia\\
\email{\{umer.ramzan, 201980053\}@gift.edu.pk, \textsuperscript{*}muhd\_ehsanrana@apu.edu.my, \{usmanali, 201980033, 201980015, 201980011\}@gift.edu.pk}
}

\maketitle

\begin{abstract}
This research paper addresses the significant challenge of accurately estimating poverty levels using deep learning, particularly in developing regions where traditional methods like household surveys are often costly, infrequent, and quickly become outdated. To address these issues, we propose a state-of-the-art Convolutional Neural Network (CNN) architecture, extending the ResNet50 model by incorporating a Gated-Attention Feature-Fusion Module (GAFM). Our architecture is designed to improve the model's ability to capture and combine both global and local features from satellite images, leading to more accurate poverty estimates. The model achieves a 75\%\ $R^2$ score, significantly outperforming existing leading methods in poverty mapping. This improvement is due to the model's capacity to focus on and refine the most relevant features, filtering out unnecessary data, which makes it a powerful tool for remote sensing and poverty estimation.
\end{abstract}

\keywords{Gated-Attention Feature-Fusion Module, GFFM, SE-blocks, Auxilory layer, Poverty Estimation, Satellite Imagery}

\section{Introduction}
Poverty remains a pervasive issue that hinders the progress of human society. The United Nations 2030 Agenda for Sustainable Development underscores the importance of addressing poverty by making it the first of its 17 Sustainable Development Goals (SDGs) - to end poverty in all its forms everywhere \cite{ref1}. Despite ongoing research and efforts, accurately assessing poverty levels continues to be a significant challenge, especially in developing countries where resource constraints and logistical difficulties impede data collection. Traditional methods, such as household surveys, although valuable, are labor-intensive, costly, and infrequent. This results in data that can quickly become outdated and lack comprehensive coverage. As a result, the lack of accurate and timely data severely limits the ability of governments and aid organizations to effectively target and allocate resources to those who need them most \cite{ref2}.

In contrast, satellite imagery provides a readily available and up-to-date source of data that can help address these challenges.  Satellite images provide a wealth of information about various physical and environmental features that correlate with economic activity \cite{ref3}. Similarly, nightlights data, which reflects the brightness of areas at night, serves as a proxy for economic activity and infrastructure development \cite{ref4}. The intensity of these nightlights can indicate levels of development, making it crucial for estimating poverty. By integrating satellite and nightlight data, we can use techniques like transfer learning to gain deeper insights into poverty distribution. This combined approach not only enhances the accuracy of poverty mapping but also enables more timely and effective interventions by taking advantage of the strengths of both data sources.

The pioneering work by Jean \textit{et al.} \cite{ref5} marked a significant milestone by introducing the use of satellite imagery combined with machine learning to estimate poverty levels. They employed Convolutional Neural Networks (CNNs) to predict nighttime light intensities based on daytime images, creating a method to infer economic activity and wealth from satellite data. This innovative approach laid the foundation for further research in the field, which has seen considerable advancements over time. Notably, the application of ensemble learning techniques, which merge models like ResNet variants, has enhanced the ability to minimize spatial noise and make more accurate poverty predictions on a detailed scale \cite{ref6}. A study that utilized Open Street Maps and land cover data from the European Space Agency (ESA) to estimate multidimensional poverty indices found that the Random Forest algorithm was the top performer among various machine learning methods \cite{ref7}. Moreover, the use of Recurrent Neural Networks (RNNs) to process satellite images has shown potential in predicting wealth indices and pinpointing poverty hotspots \cite{ref8}. 

In this paper, we proposed a state-of-the-art GAFM-based framework using transfer learning \cite{ref9} for the poverty prediction task. In our framework, the CNN is trained to learn relevant features from daytime satellite imagery to predict nightlight intensities. The GAFM module captures broad, global features and local details, fusing them effectively to ensure a balanced contribution of both local and global features that results in information-rich representations. This fusion process significantly improves the accuracy of nightlight intensity prediction, which is crucial for the subsequent income prediction task, thereby establishing the GAFM framework as a powerful tool for poverty assessment.

Following the introduction, Section \ref{sec:related_work} reviews related work in poverty prediction using satellite imagery, defining the context and importance of our contributions. Section \ref{sec:method} details the methodology, including the architecture of the Gated-Attention Feature-Fusion module. Section \ref{sec:results and discussion} presents the experimental results through quantitative analysis. Section \ref{sec:Conclusion} concludes by summarising the key contributions and discussing the impact of our work on the field of poverty estimation using satellite imagery.

\section{Related Work}\label{sec:related_work}
Deep learning has transformed various fields, enabling tasks such as disease diagnosis from medical images \cite{ref10}, fraud detection in finance \cite{ref11}, and the enhancement of autonomous systems \cite{ref12}. In remote sensing, Convolutional Neural Networks (CNNs) have been pivotal in analyzing satellite imagery for environmental monitoring, urban planning, and disaster response \cite{ref13}, with applications extending to socio-economic predictions, including poverty estimation. By leveraging satellite imagery, researchers have developed models that infer economic indicators from environmental features like infrastructure quality and agricultural activity \cite{ref14}, offering scalable and cost-effective alternatives to traditional survey-based methods. 

To improve the accuracy and efficiency of poverty prediction, various machine learning techniques have been explored. For instance, studies comparing econometric models and machine learning methods, such as logistic regression and random forest, have demonstrated the robustness of machine learning, particularly in diverse conditions \cite{ref15}. Ensemble methods have also been effective, utilizing variables like population density and nighttime lights to achieve high prediction accuracy \cite{ref16}. In Nigeria, techniques like LASSO and boosted regressions have shown promising results with AUC values between 0.79 and 0.85 \cite{ref17}, while non-parametric approaches, such as Gaussian Process regression and elastic net regularization, have exhibited strong performance in predicting variables related to health and living standards \cite{ref18}.

Jean \textit{et al.} \cite{ref5} introduced the use of machine learning techniques applied to satellite imagery for poverty prediction. They utilized a convolutional neural network (CNN) to predict nighttime light intensities from daytime images as proxies for economic activity, followed by ridge regression to estimate consumption expenditure and asset wealth. Subsequent studies, such as by Yeh \textit{et al.} \cite{ref19}, continued using nightlights as proxies but varied in their modeling and data approaches. However, these methods have limitations in regions with low luminosity, which correlate poorly with extreme poverty levels \cite{ref5}. To address these limitations, researchers have explored additional economic indicators like building density, road characteristics, roof types, and sources of lighting and drinking water \cite{ref14}. These studies, while promising, face challenges in scalability due to reliance on disparate proprietary datasets. Further work has leveraged publicly available survey and satellite imagery data, expanding on the pioneer's methodology \cite{ref5} with models like ResNet-18 trained on Landsat and nightlight data, and using parallel model training and concatenation techniques for improved predictions \cite{ref19}. In regions such as Nigeria and Rwanda, CNN models like ResNet-101 and VGG16, respectively, have been utilized for poverty mapping, with notable success in extracting features and predicting wealth indices using both daytime and nighttime imagery gaining an $R^2$ of 0.496 \cite{ref20}.

Recent research has focused on ensemble learning techniques to enhance robustness and predictive accuracy. Agyemang \textit{et al.} \cite{ref6} used a transfer learning approach with an ensemble of CNN models to predict chronic poverty in rural Sindh, Pakistan, outperforming previous models. Muñetón \textit{et al.} \cite{ref7} employed algorithms like Catboost, Lightboost, and Random Forest, utilizing data from OpenStreetMap and ESA land cover, to estimate multidimensional poverty indices in Medellín, Colombia, with Random Forest showing superior performance. Similarly, Puttanapong \textit{et al.} \cite{ref16} integrated geospatial datasets to predict poverty distribution in Thailand, highlighting Random Forest's effectiveness. Additionally, RNNs have been used to analyze satellite images for wealth forecasting \cite{ref21}, and Rekha \textit{et al.} \cite{ref8} employed DNNs and GANs to enhance poverty prediction by generating synthetic satellite images.
Our contribution in this work is the development of a novel Gated-Attention Feature-Fusion Module (GAFM) framework that enhances poverty estimation from satellite imagery. By extending ResNet50 model with an Auxiliary Branch, GFFM, and SE-blocks, our approach integrates both global and local features for more accurate poverty predictions. This framework improves predictive accuracy, demonstrated by a significant $R^2$ score increase, establishing GAFM as a robust tool for poverty assessment in remote sensing.
\section{Methodology}\label{sec:method}
For our research on poverty estimation via satellite imagery, several types of datasets are essential. Primarily, we require high-resolution satellite imagery and nightlight data to capture physical and economic indicators from space. Additionally, survey data is necessary to serve as ground truth for our models, providing accurate information on socioeconomic conditions.

\subsection{Data Collection}
We utilized satellite imagery sourced from Google Earth \cite{ref22}, providing detailed visual data on land use, infrastructure, and environmental features. A collection of 2,596 images at a resolution of 720x500 pixels were sourced across various districts of Punjab Pakistan. We sourced the nightlight data from NASA Earth Observatory's Black Marble, specifically the BlackMarble 2016 geogray dataset \cite{ref23}. We utilized the C1 GeoTIFF tile, as it includes Pakistan, aligning with our focus area. Each pixel in the GeoTiFF image grid corresponds to a 500-meter square on the Earth's surface. Survey data complements satellite imagery by providing ground truth for poverty estimation, validating our models, and offering a comprehensive view of socioeconomic conditions. The Pakistan Bureau of Statistics (PBS) offers relevant datasets, including demographic statistics, labor force data, agricultural statistics, and business registers. Among these surveys, the Pakistan Social and Living Standards Measurement (PSLM) survey, especially the PSLM-2020 income dataset, is critical for our research \cite{ref24}. It provides detailed information on household earnings from various sources, offering a granular view of income distribution essential for poverty estimation.

\subsection{Data Preparation}
Given the diverse nature of our datasets, we undertook several preprocessing steps to ensure the data was suitable for analysis and model training. For the survey dataset, we extracted relevant fields from the survey dataset, covering various income sources such as primary and secondary job incomes, rent, pensions, and remittances. The data was aggregated from individual to household level. While most income figures were annual, primary job income was recorded monthly only; missing annual values were extrapolated to annual level. Finally, all sources of annual income were aggregated into a single field. The aggregated annual household income was scaled into monthly and subsequently daily figures. Since satellite images cover extensive areas, we organized the data into clusters of households, aggregating income for each cluster and then averaging across the households, resulting in per cluster, per day, per house income thus giving our desired wealth indicator. We utilized district-level data for geographic locations. For each coordinate, the average nightlight intensity was extracted to quantify the luminosity of the corresponding cluster. Leveraging established research that correlates nightlight intensity with economic activity \cite{ref4}, we inferred economic status from the nightlight data \cite{ref5}. Specifically, clusters were categorized by income levels, with nightlight intensities assigned accordingly: lower intensities for low-income clusters and higher intensities for high-income clusters. Each acquired image represents a cluster, with nightlight intensity data reflecting the presumed economic activity of the area.

\subsection{Poverty Prediction Framework}

\begin{figure*}[http]
\centering
\includegraphics[width=0.8\textwidth]{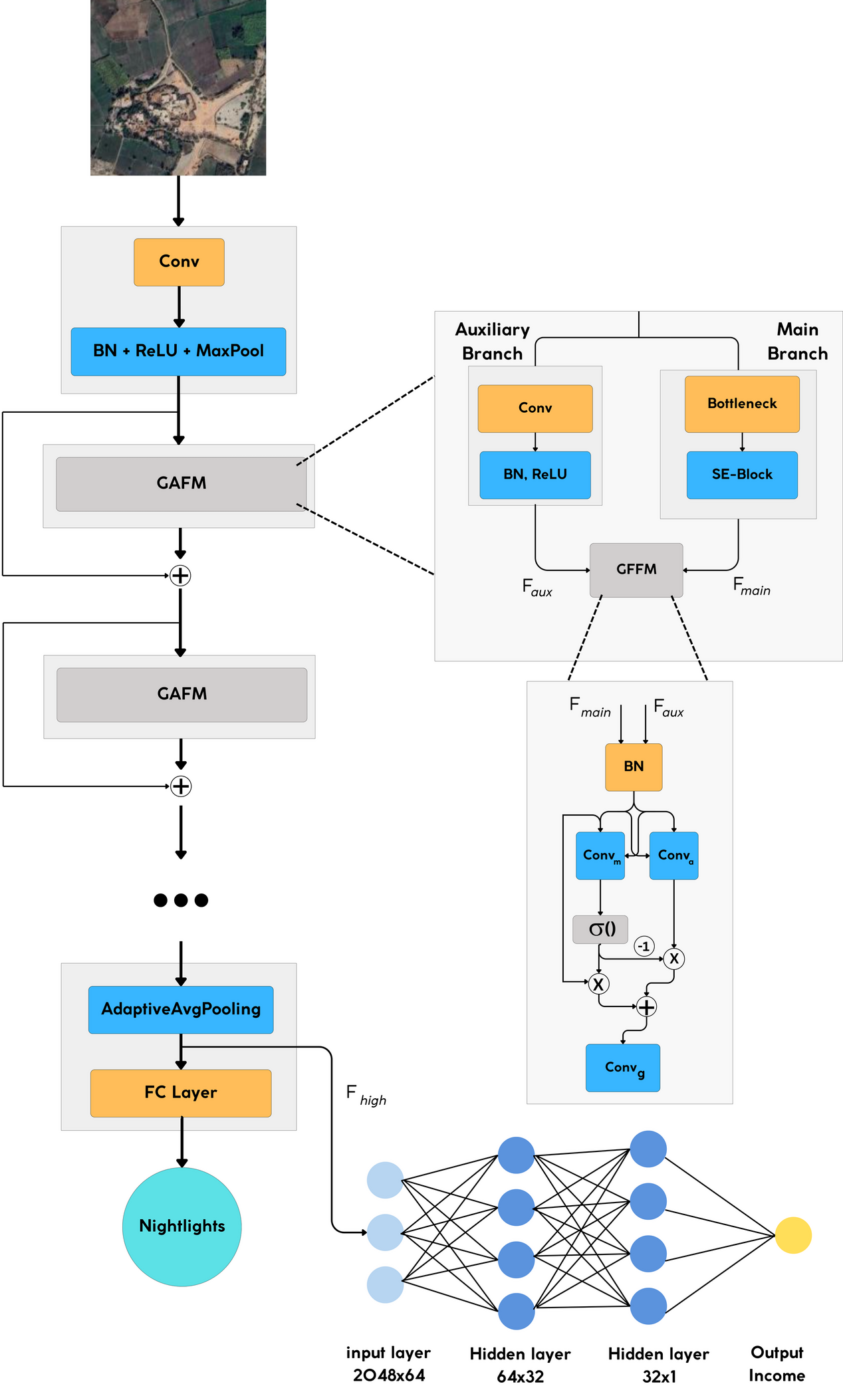}
\caption{Illustration of the GAFM framework.}
\label{fig:architecture}
\end{figure*}

We propose a novel state-of-the-art CNN architecture for poverty estimation task, as illustrated in Figure \ref{fig:architecture}. Our architecture is extending the ResNet50 model by incorporating Gated-Attention Feature-Fusion Module (GAFM) which includes an Auxiliary Branch, Gated Feature Fusion Module (GFFM), and Squeeze-and-Excitation (SE) blocks. These enhancements are designed to improve the model’s ability to estimate poverty levels from satellite images, leveraging both local and global feature representations. We apply a transfer learning approach structuring the flow into two phases: nightlights prediction serving as a proxy for economic activity and then income prediction.

The architecture begins with the input images \(\mathbf{X} \in \mathbb{R}^{H \times W \times C}\), where \(H\), \(W\), and \(C\) represent the height, width, and channels of the image, respectively, passing through initial convolutional layers to extract basic features. Let \(\mathbf{F}_{\text{conv}}\) denote the feature maps obtained after these initial convolutions. 
Conventional CNN architectures process the entire image uniformly, generating feature maps where each channel captures distinct aspects of the visual input, inherently treating all feature maps with equal importance. Given the nature of our task, architectural elements like buildings, roads, and other infrastructure have a stronger correlation with an area's economic status \cite{ref5}. Thus, it is crucial to allocate additional attention to these features, ensuring a more targeted and relevant feature extraction aligned with our objectives. To address this, we employed the GAFM block where the input is split into two paths, one leading to the Auxiliary Branch and the other to the Main Branch. The Auxiliary Branch retains coarser, global features, providing a broad context. The feature maps \(\mathbf{F}_{\text{aux}}\) from the Auxiliary Branch are computed through a series of operations as follows:

\begin{equation}
\mathbf{F}_{\text{aux}} = \text{ReLU}\left(\text{BN}\left(\text{Conv}\left(\mathbf{F}_{\text{res}}, \mathbf{W}_{\text{aux}}\right)\right)\right)
\end{equation}

Here, \(\text{Conv}(\mathbf{F}_{\text{conv}}, \mathbf{W}_{\text{aux}})\) represents the convolution operation applied to the input feature maps \(\mathbf{F}_{\text{conv}}\) using the weights \(\mathbf{W}_{\text{aux}}\), and \(\text{BN}(\cdot)\) denotes the batch normalization operation that standardizes the activations. Finally, the \(\text{ReLU}(\cdot)\) activation function introduces non-linearity into the model, resulting in the output \(\mathbf{F}_{\text{aux}}\). Meanwhile, the Main Branch employs SE-Bottleneck layers. The SE blocks within these bottleneck layers serve as an attention mechanism \cite{ref25}, selectively enhancing the most relevant feature maps while suppressing the irrelevant ones. Let \(\mathbf{F}_{\text{b}}\) denote the feature maps obtained after the bottleneck layer. For the output feature maps \(\mathbf{F}_{\text{main}}\) from the Main Branch, the SE block can be described as:

\begin{equation}
\mathbf{F}_{\text{main}} = \sigma\left(\mathbf{W}_2 \cdot \text{ReLU}\left(\mathbf{W}_1 \cdot \text{GAP}(\mathbf{F}_{\text{b}})\right)\right) \cdot \mathbf{F}_{\text{b}}    
\end{equation}

where \(\text{GAP}(\cdot)\) denotes global average pooling, \(\mathbf{W}_1\) and \(\mathbf{W}_2\) are learnable weights, and \(\sigma(\cdot)\) is the sigmoid activation function. The SE block recalibrates the channel-wise feature responses, ensuring that only the most informative features are emphasized. Once the relevant features have been identified and redundant ones suppressed, the GFFM is applied. Proposed by Park and Shin \cite{ref26}, the GFFM fuses the global features from the Auxiliary Branch with the refined local features from the Main Branch. The output feature maps \(\mathbf{F}_{\text{gffm}}\) are computed as follows:

\begin{equation}
    \mathbf{F}_{\text{gffm}} = \mathbf{W}_g \cdot (f_m \otimes \mathbf|{F}_{\text{main}}| + (1 - f_m) \otimes \mathbf{W}_a \cdot \left(\mathbf|{F}_{\text{aux}}| \odot \mathbf|{F}_{\text{main}}|\right)
\end{equation}

here, 
\begin{equation}
f_m = \sigma \left( \mathbf{W}_m \cdot \left( \mathbf|{F}_{\text{main}}| \odot \mathbf|{F}_{\text{aux}}| \right) \right)
\end{equation}

where \(\odot\) and \(\otimes\) denotes concatenation and element-wise multiplication, \(\mathbf{W}_g(\cdot)\), \(\mathbf{W}_a(\cdot)\) and \(\mathbf{W}_m(\cdot)\) are convolution operations, representing the learnable gating weights. The gating weights modulate the contribution of each feature map, allowing the model to dynamically adjust the balance between global and local features based on the specific context of the input data. This process ensures that the fused feature maps are both rich and focused, optimizing the network’s ability to predict nightlight intensities. Finally, the features obtained from GAFM blocks are given to a fully connected layer, defined as:

\begin{equation}
\hat{I} = \text{FC}\left(\text{AvgPool}\left(\mathbf{F}_{\text{GAFM}}\right)\right)
\end{equation}
where \(\mathbf{F}_{\text{GAFM}}\) is the output from the last GAFM block, \(\text{AvgPool}(\cdot)\) denotes the adaptive average pooling operation, and \(\text{FC}(\cdot)\) represents the fully connected layer that produces the final output $\hat{I}$ i.e, the nightlight intensity.

During this phase, we employed the Mean Squared Error (MSE) loss function to guide the optimization process, defined as:

\begin{equation}
\mathcal{L}_{\text{MSE}} = \frac{1}{N} \sum_{i=1}^{N} \left(I_i - \hat{I}_i\right)^2    
\end{equation}

where \(y_i\) and \(\hat{y}_i\) represent the ground truth and predicted nightlight intensities, respectively, and \(N\) is the number of samples. After the training process, the last layer is removed, and the CNN is transformed into a feature extractor. This modified CNN then generates high-level representations of the satellite images, let it me denoted as \(\mathbf{F}_{\text{high}}\). These high-level features are then fed into a Vanilla Neural Network (VNN) for income prediction, which can be described as:

\begin{equation}
\hat{y} = \mathbf{W}_3 \cdot \text{ReLU}\left(\mathbf{W}_2 \cdot \text{ReLU}\left(\mathbf{W}_1 \cdot \mathbf{F}_{\text{high}} + \mathbf{b}_1\right) + \mathbf{b}_2\right) + \mathbf{b}_3    
\end{equation}

where \(\mathbf{W}_1\), \(\mathbf{W}_2\), and \(\mathbf{W}_3\) are the weight matrices, \(\mathbf{b}_1\), \(\mathbf{b}_2\), and \(\mathbf{b}_3\) are the bias vectors, and \(\hat{y}\) represents the predicted income levels. With the goal of minimizing the absolute difference between the predicted and actual income levels, we utilize the Mean Absolute Error (MAE) loss function defined as:

\begin{equation}
\mathcal{L}_{\text{MAE}} = \frac{1}{N} \sum_{i=1}^{N} \left| y_i - \hat{y}_i \right|    
\end{equation}

where \(y_i\) and \(\hat{y}_i\) represent the ground truth and predicted income levels, respectively. The VNN leverages the refined features by taking advantage of the attention mechanism within the SE blocks, the broader contextual insights provided by the Auxiliary Branch, and the selective gating of the GFFM. This combination allows the VNN to accurately translate features from satellite imagery into income levels, making it a powerful tool for estimating poverty.

This architecture, by focusing on filtering and refining the feature maps through SE blocks and GFFM, effectively adapts the model to the unique demands of satellite imagery analysis. It preserves both global and local information while ensuring that only the most relevant features are passed forward, leading to a precise and efficient model for remote sensing applications in poverty estimation.

\section{Experimental Results}\label{sec:results and discussion}

In our poverty estimation approach, the model was initially trained using the Adam optimizer with a learning rate of 0.001, alongside the \texttt{ReduceLROnPlateau} scheduler, which dynamically adjusted the learning rate during training. The training was conducted on hardware including an NVIDIA GeForce RTX 4080 GPU with 16GB of dedicated memory, 64GB RAM, and an Intel(R) Xeon(R) CPU E5-2667 v3 @ 3.20GHz. The first phase involved 100 epochs of training over approximately 1.2 hours, while the second phase, consisting of 50 epochs, took about 1 minute. The model's performance was evaluated using k-fold cross-validation with the $R^2$ metric. The $R^2$ score is defined as:

\begin{equation}
R^2 = 1 - \frac{\sum_{i=1}^{n} (y_i - \hat{y}_i)^2}{\sum_{i=1}^{n} (y_i - \bar{y})^2}
\end{equation}

where \( y_i \) represents the actual values, \( \hat{y}_i \) represents the predicted values, and \( \bar{y} \) is the mean of the actual values.

\begin{figure}[h]
    \centering
    \begin{tabular}{cc}
        \includegraphics[width=0.45\linewidth]{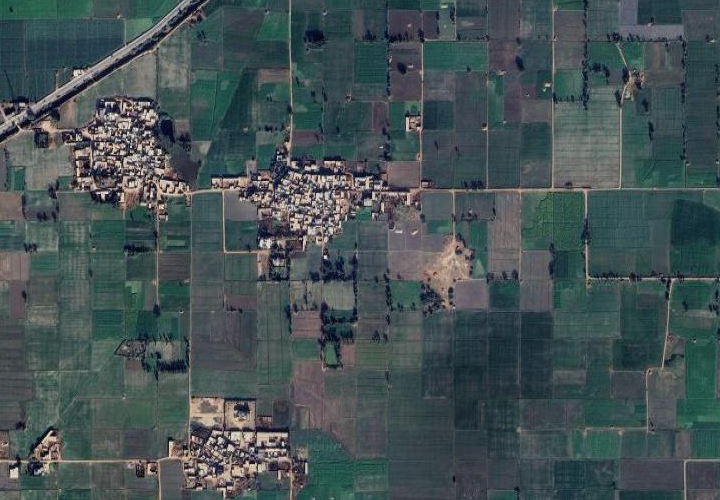} &
        \includegraphics[width=0.45\linewidth]{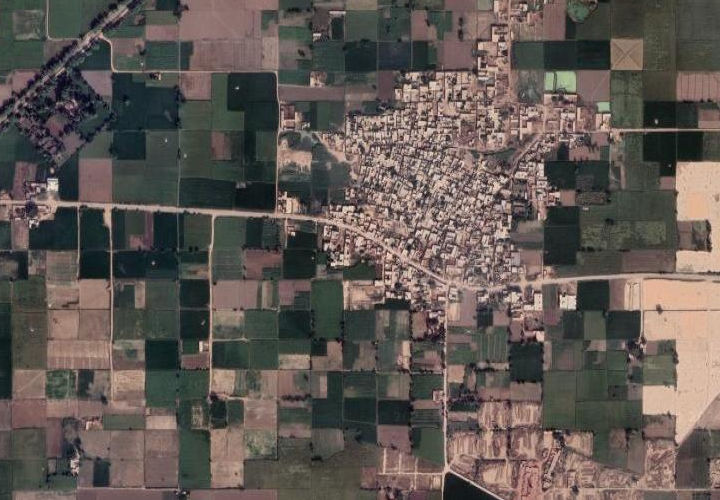} \\
        (a) & (b) \\
        \includegraphics[width=0.45\linewidth]{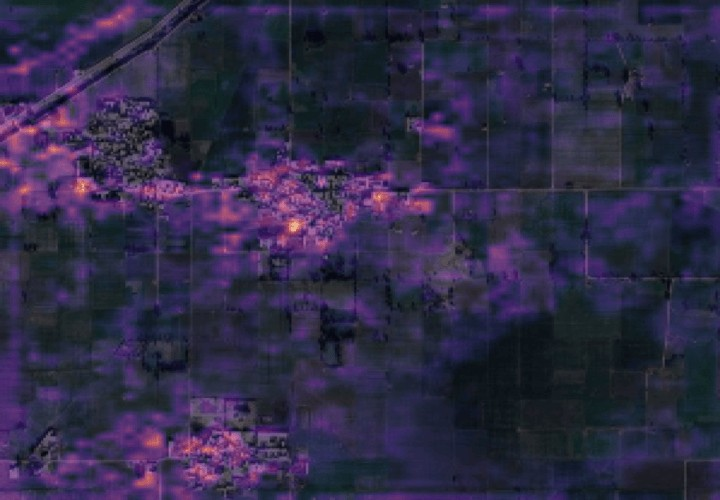} &
        \includegraphics[width=0.45\linewidth]{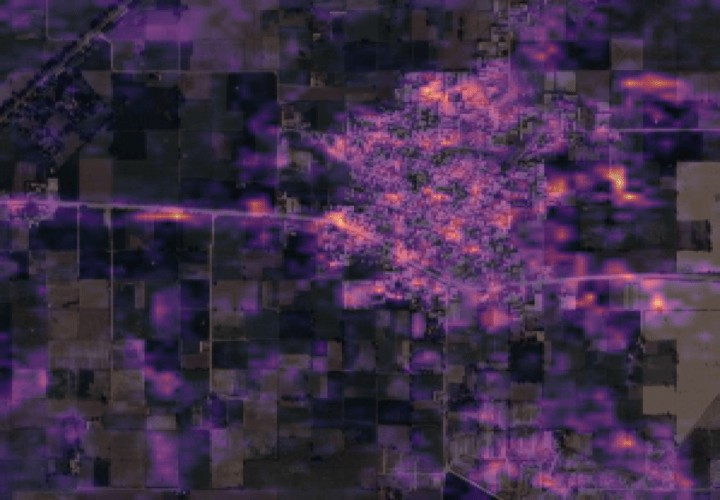} \\
        (c) & (d) \\
    \end{tabular}
    \caption{Illustration of input images alongside their corresponding activation maps, demonstrating the model's focus on salient features during the prediction process.}
    \label{fig:sample_sat_imgs}
\end{figure}

To thoroughly evaluate our proposed methodology, we applied several state-of-the-art techniques to our dataset. Our methodology achieved a 75\% $R^2$ score, outperforming these approaches. Specifically, it demonstrated a 74\% improvement over Jean \textit{et al.} \cite{ref5}, a 55\% improvement over Li \textit{et al.} \cite{ref27}, and a 60\% improvement over Rekha \textit{et al.} \cite{ref28} in terms of the $R^2$ metric. While our approach shows substantial advancements, it is important to acknowledge the innovative contributions of these studies. For instance, Rekha \textit{et al.} \cite{ref28} enhanced nightlight intensity estimation by incorporating both daytime and nighttime images, providing a more comprehensive understanding of urban environments. Additionally, Li \textit{et al.} \cite{ref27} methodology applied advanced techniques such as LSD, Hough Transform, GLCM, LBP, and HOG, focusing on structural and textural feature extraction from remote sensing imagery. These techniques underscored the significance of textural features in identifying urban poverty. However, unlike these methods that focus solely on either global or local features and treat them equally, our proposed GAFM-framework effectively combines both, ensuring that each type of feature is selectively and strategically utilized to enhance the model’s predictions. Our methodology not only identifies key poverty predictors, such as infrastructure, but also filters out irrelevant information like barren lands, as demonstrated in Figure \ref{fig:sample_sat_imgs}. This targeted approach allows our model to focus on the most informative aspects of the satellite images, leading to a significant improvement in poverty estimation accuracy.

\section{Conclusion}\label{sec:Conclusion}
In this study, we demonstrated the efficacy of our state-of-the-art Gated-Attention Feature-Fusion Module (GAFM) framework for poverty estimation. The GAFM framework outperformed existing methods, achieving an $R^2$ of 75\%  thus establishing itself as a powerful tool for identifying poverty. Its precision in pinpointing impoverished areas underscores its potential in poverty assessment. The results indicate that our framework enhances the accuracy and timeliness of poverty identification, thereby supporting the development of more effective policies for poverty reduction. Further optimization of the GAFM-framework can be achieved by fine-tuning its parameters and exploring alternative layers and architectures to improve accuracy and efficiency. Moreover, applying the GAFM framework to diverse geographical regions with varying cultural and economic contexts will provide valuable insights into its performance, allowing for a comprehensive assessment of its generalizability and adaptability across different environments.

\end{document}